\documentclass[letterpaper, 10 pt, journal, twoside]{ieeetran}
\usepackage[left=16.9mm,right=16.9mm,top=20.1mm,bottom=15.2mm,footskip=5mm]{geometry}

\usepackage{amsmath,amsfonts}
\usepackage{algorithmic}
\usepackage{algorithm}
\usepackage{array}
\usepackage{graphicx}
\usepackage[caption=false,font=normalsize,labelfont=sf,textfont=sf]{subfig}
\usepackage{textcomp}
\usepackage{stfloats}
\usepackage{url}
\usepackage{verbatim}
\usepackage{graphicx}
\usepackage{cite}
\usepackage{xcolor}
\usepackage{soul}
\usepackage{svg}
\usepackage{multirow}
\usepackage{hyperref}
\usepackage{siunitx}
\usepackage{soul}
\usepackage{fancyhdr}

\fancypagestyle{firststyle}
{
   \fancyfoot[C]{\footnotesize	{\color{red}\textbf{© 2022 IEEE.  Personal use of this material is permitted.  Permission from IEEE must be obtained for all other uses, in any current or future media, including reprinting/republishing this material for advertising or promotional purposes, creating new collective works, for resale or redistribution to servers or lists, or reuse of any copyrighted component of this work in other works.}}}
   \lhead{\color{red}\textbf{PREPRINT VERSION}}%
   \rhead{\color{red}\textbf{FINAL VERSION AT IEEE Xplore® with DOI: 10.1109/LRA.2022.3219306}}%
}

\begin{document}

\title{Hardware-accelerated Mars Sample Localization via deep transfer learning from photorealistic simulations}

\author{R. Castilla-Arquillo$^{1}$, C. J. Pérez-del-Pulgar$^{1}$, G. J. Paz-Delgado$^{1}$ and L. Gerdes$^{2}$%
\thanks{This work has been partially funded by the Andalusian Regional Government under the project entitled "Intelligent Multimodal Sensor for Identification of Terramechanic Characteristics in Off-Road Vehicles (IMSITER)" under grant agreement P18-RT-991.} 
\thanks{$^{1}$R. Castilla-Arquillo, C. J. Pérez-del-Pulgar and G. J. Paz-Delgado are with the Department of Automation and Systems Engineering, Universidad de Málaga, Andalucía Tech, 29070 Málaga, Spain (e-mail: raulcastar@uma.es; carlosperez@uma.es; gonzalopd96@uma.es).}
\thanks{$^{2}$L. Gerdes is with the Automation and Robotics Section of the European Space Agency, Keplerlaan 1, 2201 AZ Noordwijk, The Netherlands (e-mail: levin.gerdes@esa.int).}
}

\maketitle
\thispagestyle{firststyle}
\pagestyle{fancy}

\begin{abstract}
The goal of the Mars Sample Return campaign is to collect soil samples from the surface of Mars and return them to Earth for further study. The samples will be acquired and stored in metal tubes by the Perseverance rover and deposited on the Martian surface. As part of this campaign, it is expected that the Sample Fetch Rover will be in charge of localizing and gathering up to 35 sample tubes over 150 Martian sols. Autonomous capabilities are critical for the success of the overall campaign and for the Sample Fetch Rover in particular. This work proposes a novel system architecture for the autonomous detection and pose estimation of the sample tubes. For the detection stage, a Deep Neural Network and transfer learning from a synthetic dataset are proposed. The dataset is created from photorealistic 3D simulations of Martian scenarios. Additionally, the sample tubes poses are estimated using Computer Vision techniques such as contour detection and line fitting on the detected area. Finally, laboratory tests of the Sample Localization procedure are performed using the ExoMars Testing Rover on a Mars-like testbed. These tests validate the proposed approach in different hardware architectures, providing promising results related to the sample detection and pose estimation.
\end{abstract}


\section{INTRODUCTION} 
\IEEEPARstart{T}{he} Mars Sample Return (MSR) campaign consists of several missions whose main objective is the collection of Martian geological samples for their analysis on Earth. Mars 2020 is part of this campaign, where the Perseverance rover will be in charge of encapsulating scientifically relevant samples into tubes. The rover is expected to collect about 35 samples, which will be placed on the Martian surface \cite{Farley2020}. Each sample tube location will be tagged in orbital images along with on-site images taken by the Perseverance rover at the moment of sample acquisition. Retrieval of the cached sample tubes will be performed by the Sample Fetch Rover (SFR) in the Sample Retrieval Lander (SRL) mission. This rover is scheduled to traverse up to \SI{20}{\km} during the first 150 sols (Martian days) \cite{Muirhead2020}. A high degree of autonomous driving capabilities is necessary to satisfy these requirements. The SFR will be solar powered, thus presenting energy constraints which limit possible solutions. In addition, the SFR would need to rely on visual localization techniques as tagged orbital sample tubes positions would not be precise enough.

\begin{figure}[]
\centering
\includegraphics[width=0.45\textwidth]{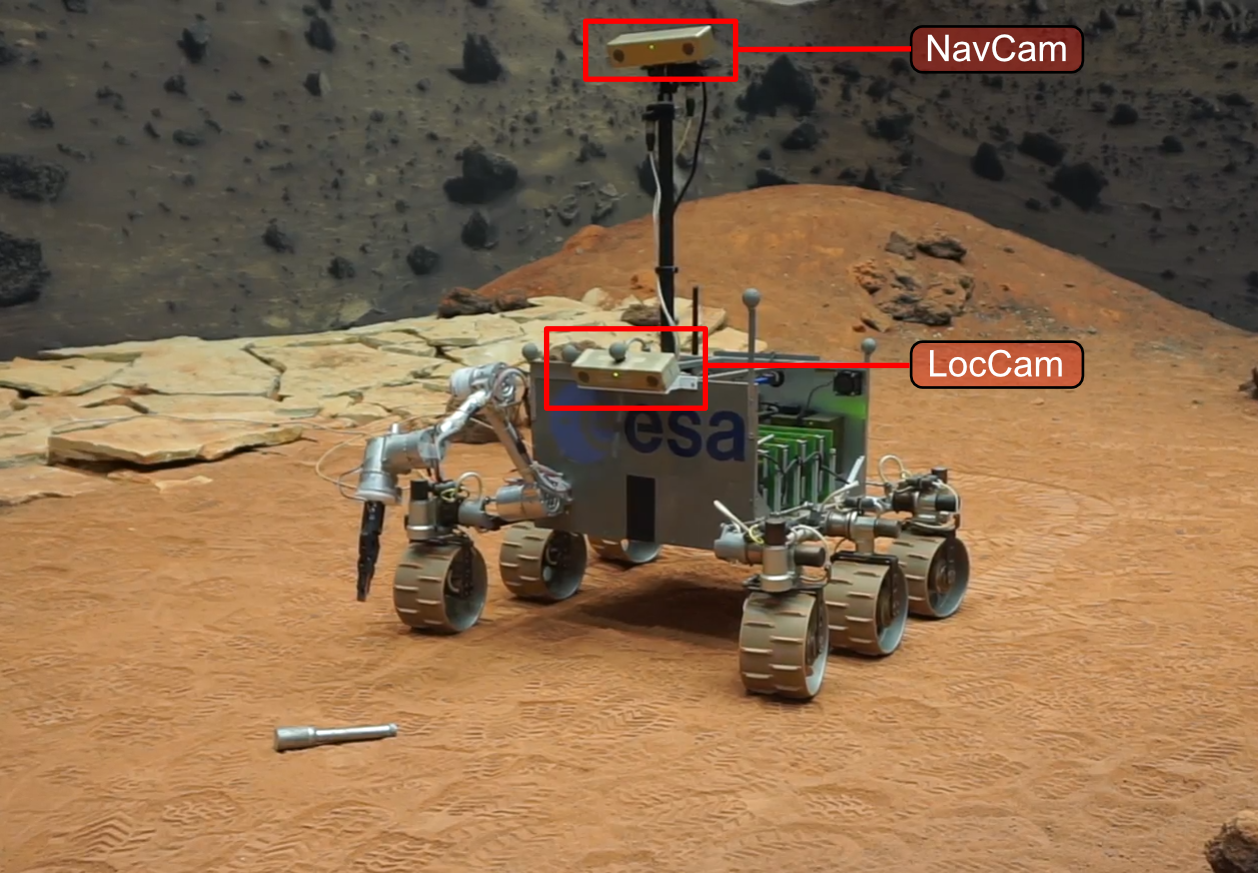}
\caption{Picture of ExoTeR \cite{azkarate2022design} taken at the Planetary Utilisation Testbed. The rover is equipped with a manipulator, a gripper and two stereo cameras: LocCam and NavCam. The sample tube is located on sandy terrain in front of it.}
\label{fig:exoter-rover}
\end{figure}

Autonomous sample tube localization with onboard cameras helps to improve the overall SFR mission. Studies have proposed to estimate the pose of the tubes using Computer Vision (CV) techniques \cite{Edelberg2015,Lee2018} and semantic image segmentation \cite{Papon2017,Daftry2021}. Deep Neural Networks (DNNs) have been proven to be more effective than humans in tasks such as image classification \cite{Alzubaidi2021}. In fact, machine learning techniques have already been applied in Mars missions \cite{Nelessen2019} for the elaboration of landing hazard maps, taking into account the density of rocks of the landing area. However, the implementation of DNNs in space-related projects poses several challenges.

The first challenge is the need of training datasets with a high volume of images to accurately detect a specific target. This is usually solved by creating large datasets of the target in a wide range of scenarios, i.e., sample tubes on different surfaces and illumination conditions. However, taking images of the target in real conditions is rather complex when it is located in an unreachable environment such as the Martian surface. There are studies that tackle this issue creating datasets based on special testbeds that try to replicate some of the characteristics presented in the real environment \cite{Pham2020}. Another approach is to use deep transfer learning techniques to take advantage of the characteristics learnt from a specific domain dataset, adapting the pre-trained parameters into a new application. DNNs have demonstrated promising results in domains with insufficient training data \cite{Tan2018,Shorten2019}. Furthermore, synthetic image generation has been used to create space-related datasets that allowed training a DNN for possible adverse conditions \cite{Proencca2020,DelgadoCenteno2021,Pham2021}.

The second challenge is related to the computational demands of DNNs. High-end computers with powerful CPUs and GPUs are traditionally adopted to execute this kind of neural networks. However, space-grade CPUs are usually more limited in performance than CPUs for terrestrial applications. Furthermore, DNNs consume considerable electrical power, a critical resource for solar powered robotic systems such as rovers. An alternative is to complement these resource-constrained embedded processors with FPGA-based DNN accelerators and/or custom, energy efficient Application Specific Integrated Circuits (ASICs) \cite{Chen2019}. It is important to choose the appropriate DNN architecture for the hardware in which it will be implemented \cite{Kothari2020}. Topologies with a reduced number of layers and interconnections can facilitate the implementation on hardware.

In this system paper, we propose a novel architecture to solve the Mars Sample Localization problem, allowing a rover to autonomously detect and estimate the pose of sample tubes on the Martian surface. As a result, we address several challenges, which are inherently related to the proposed architecture. The first one, associated to the use of DNN to detect the samples on images, whose main drawback is the lack of training datasets with a high volume of real images. It has been solved by using deep transfer learning techniques, based on synthetic images generated on a photorealistic simulator.
The second challenge is focused on the implementation of the sample pose estimation once it has been detected, based on the use of stereo-cameras and trying to minimize computational resources.
For this purpose, we have used monochrome images since the ESA space qualified cameras onboard the future SRF and Rosalind Franklin rovers are provided with this limitation.
Finally, the proposed solutions were tested on the Planetary Utilisation Testbed, using the ExoMars Testing Rover (ExoTeR) at the European Space Agency (ESA) (depicted in Fig.~\ref{fig:exoter-rover}). This allow us to create a dataset with both synthetic and real images, which were used to demonstrate the proposed solutions, deployed on different hardware configurations to analyse their performance. 

\begin{figure}[]
  \centering
  \includegraphics[width=0.48\textwidth]{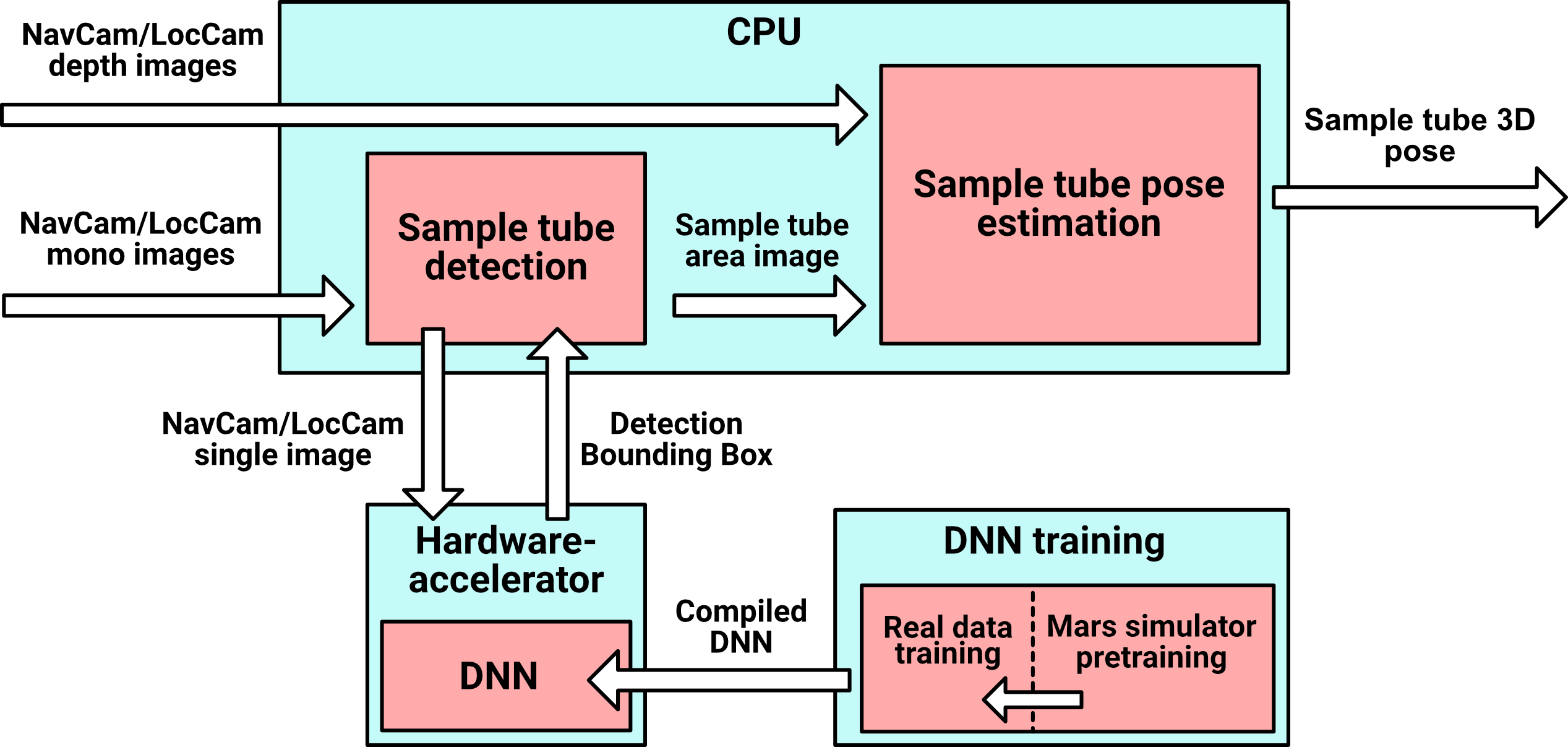}
\caption{Architecture overview of the proposed Mars Sample Localization system.}
\label{fig:system-overview}
\end{figure}

\section{SAMPLE LOCALIZATION}
Rovers are usually equipped with two stereo cameras. The first one, commonly called NavCam, is located on the rover mast and it is used for navigation purposes. The second camera, named LocCam, is located on the rover front and it is used for self localization by means of Visual Odometry methods. In addition, the LocCam is able to detect obstacles in front of the rover \cite{Gerdes2020EfficientResources}. 

The proposed Sample Localization system is divided into two main modules, which can be observed in Fig. \ref{fig:system-overview}: detection and pose estimation. In the detection module, images from both NavCam and LocCam are regularly sent to a hardware-accelerated DNN for the detection of the sample tubes. Moreover, the modularity of the proposed system makes it possible to choose between different DNNs and hardware-accelerators architectures, e.g.: FPGA-based neural networks. This network has been previously trained with both synthetic and real data to increase its accuracy in unknown environments such as the Martian surface. 
It is worth mentioning that only one sample tube model has been considered, reducing therefore the DNN size and the inference time.
Once a detection is confirmed, the network provides a bounding box of the area in which the sample tube is located which is then cropped and sent to the pose estimation module. Next, the aforementioned module provides an estimation of the sample tube 3D coordinates (position and orientation) on the terrain using the disparity images computed from both stereo cameras. For this purpose, CV techniques such as contour detection and line fitting are implemented using the detection bounding box image area.

\subsection{Detection}
\label{sec:detection}
\begin{figure*}[t]
\centering
\includegraphics[width=1\textwidth]{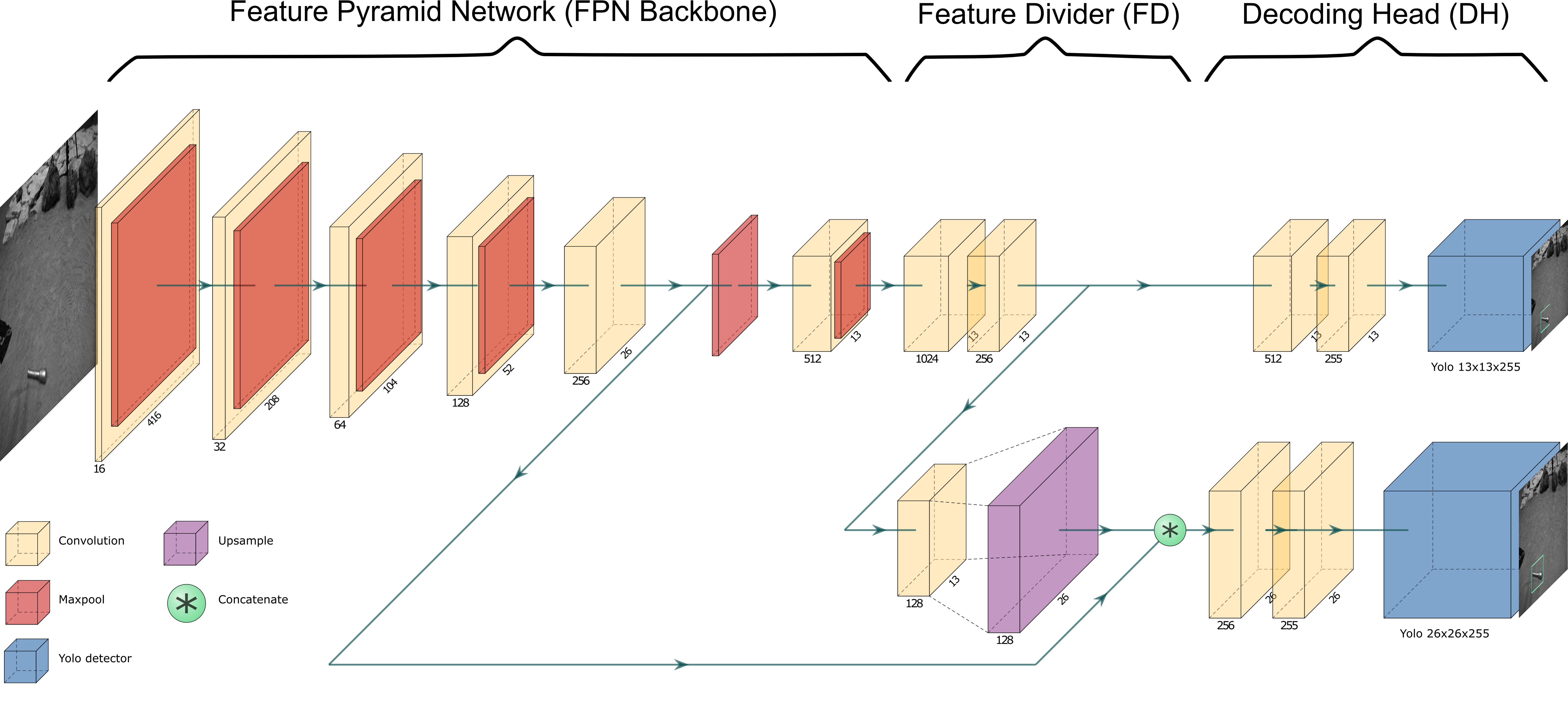}
\caption{Architecture of the YOLOv3-tiny network. It is composed by 23 layers and divided in three sectors from left to right: the Feature Pyramid Network, the Feature Divider and the Decoding Head. The FPN receives an image and process it to extracts its features. The FD combines feature maps of different level of detail and send them to the DH. Finally, the DH calculates the bounding box of the target along its likelihood. A legend of all the characteristic operational blocks can be found on the lower left corner. Numbers corresponding filters sizes can be found under each block.}
\label{fig:yolo-netwok}
\end{figure*}

A DNN is used to accomplish the detection task. Thus, the first step of this section is to train the network with images of the target to be detected. For this purpose, YOLOv3-tiny network architecture, a less dense version of YOLOv3 \cite{Redmon2018}, has been implemented. Its Backbone is based on the Darknet architecture, which is, in turn, inspired by the concept of Feature Pyramid Network (FPN) \cite{lin2017feature}. It can be inferred from Fig.~\ref{fig:yolo-netwok}. Starting from the left, an image of the sample tube is introduced into the network, which creates features maps of decreasing resolution using the FPN architecture. Later on, the Feature Divider is made up of two branches: the upper branch, focused on extracting a fine-grained feature map, and the lower branch, centered on obtaining the overall feature map. This is done by concatenating the outputs of different sections of the FPN. Finally, in the Decoding Head (DH), preliminary detections are performed for both feature maps (YOLO blocks) and fused into a final detection in the form of a bounding box of the object of interest, which will be used for pose estimation. 

\begin{figure}[]
  \centering
  \includegraphics[width=0.48\textwidth]{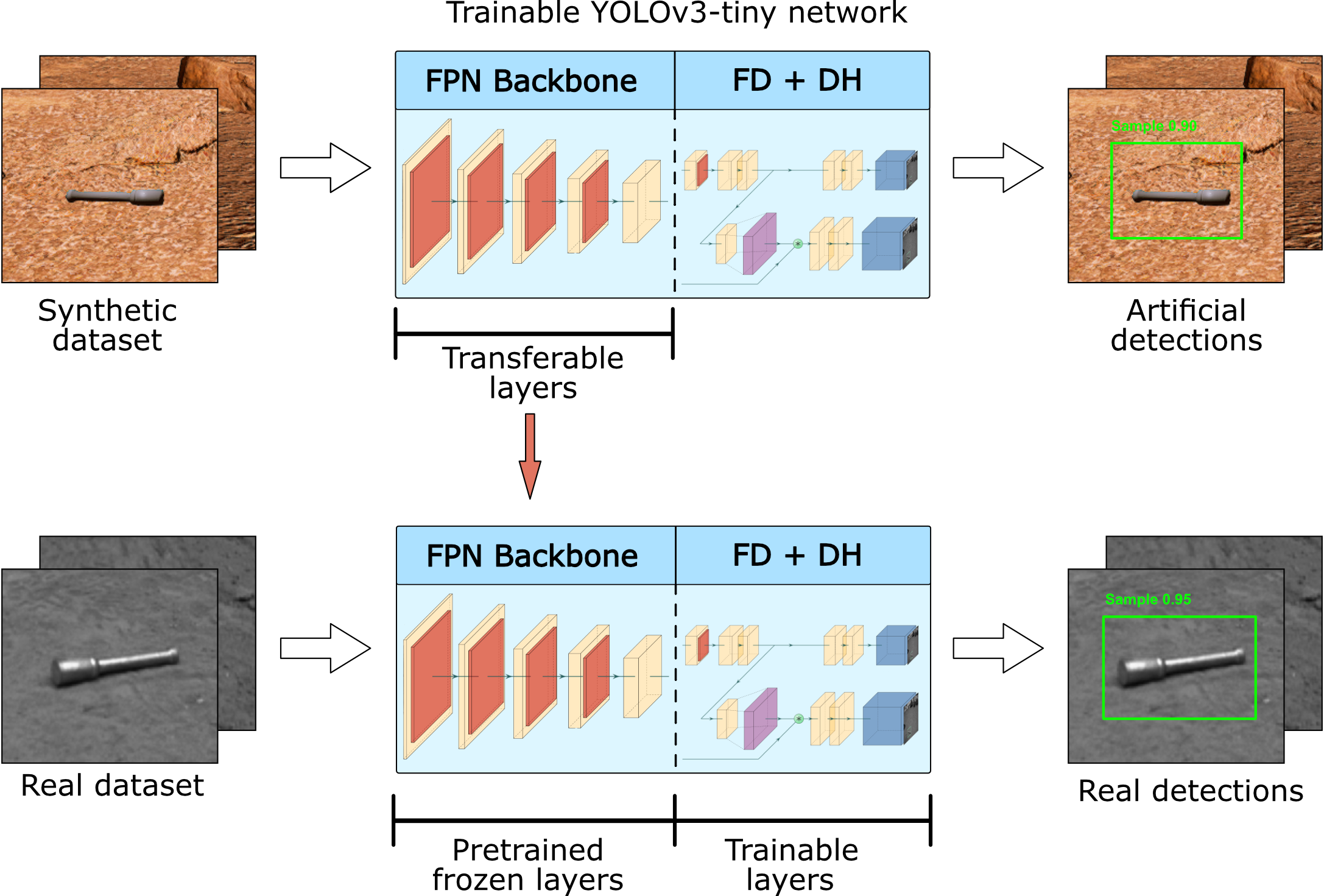}
\caption{Transfer learning technique applied to enhance the YOLOv3-tiny network detection capabilities. First, the network is trained with synthetically generated images. Later, the pre-trained first layers of the network are transferred to a second network, which will be trained with real images to perform the intended task.}
\label{fig:transfer-network}
\end{figure}

The main difference of YOLOv3-tiny over other classical network architectures is that it does not present a dense classifier network as its Decoding Head. Instead, inside the YOLO block, images are divided into regions of pre-calculated weighted bounding boxes and computed probabilities of each detection are used to choose a final bounding box. This characteristic makes the neural network faster and less dense, hence, a network more suitable for real-time applications on embedded devices.

Ultimately, a DNN performance is highly dependant on the volume of training data. Therefore, DNNs trained with large datasets usually produce better detection results that networks trained with less images. However, for the sample tube detection problem, it is rather difficult to produce a large enough dataset with real life images replicating all the possible scenarios of a Martian environment. To solve it, a synthetic dataset is generated that contains artificial images of the sample tube. These images recreate a Martian scenario using a photorealistic simulator based on Unreal Engine. They depict the sample tube in a wide range of layouts and illumination conditions, providing images as close to reality as possible. Additionally, the synthetic dataset is complemented with real life images of the sample tube on a Mars-like tesbed.

Deep transfer learning is applied for training the YOLOv3-tiny network to take full advantage of the features of the synthetic dataset. The employed method is divided into two steps, showed at Fig. \ref{fig:transfer-network}. First, a full YOLOV3-tiny model is completely trained with synthetic images of the sample tube (upper part of Fig. \ref{fig:transfer-network}). Once trained, its FPN Backbone has its layers frozen and is transferred to a second model. Like this, the previously learned feature maps (trained with a large and diverse synthetic dataset) are adapted to a new model. This second model would be the one implemented on the rover (lower part of Fig. \ref{fig:transfer-network}). This second network is trained with real images: the Feature Divider and Decoding Head sectors are trained while the FPN Backbone keeps frozen. Hence, the network specializes for real life detections, though keeping the flexibility achieved on the first model trained with the synthetic dataset. In this case, a three-channel image is introduced into the network by stacking the one-channel grayscale image and performing zero-padding while rescaling the image to the network resolution. Finally, thanks to its reduced size, the YOLOv3-tiny model is easily ported to an embedded hardware accelerator, following a process of quantization and compression of all its layers.

\subsection{Pose estimation}
In this module, the sample tube area image obtained from the detection module is used to estimate the sample tube pose in the world space coordinates. The sample tube pose can be estimated at any time of the Sample Localization process thanks to the 3D information provided by the stereo-cameras. However, this module acquires relevance when the sample tube has been detected in front of the LocCam and is reachable by the rover manipulator. Then, the algorithm can provide the required accuracy for the rover to compute the manipulator trajectory to effectively pick up the sample tube. 

\begin{figure}
  \centering
  \includegraphics[width=0.48\textwidth]{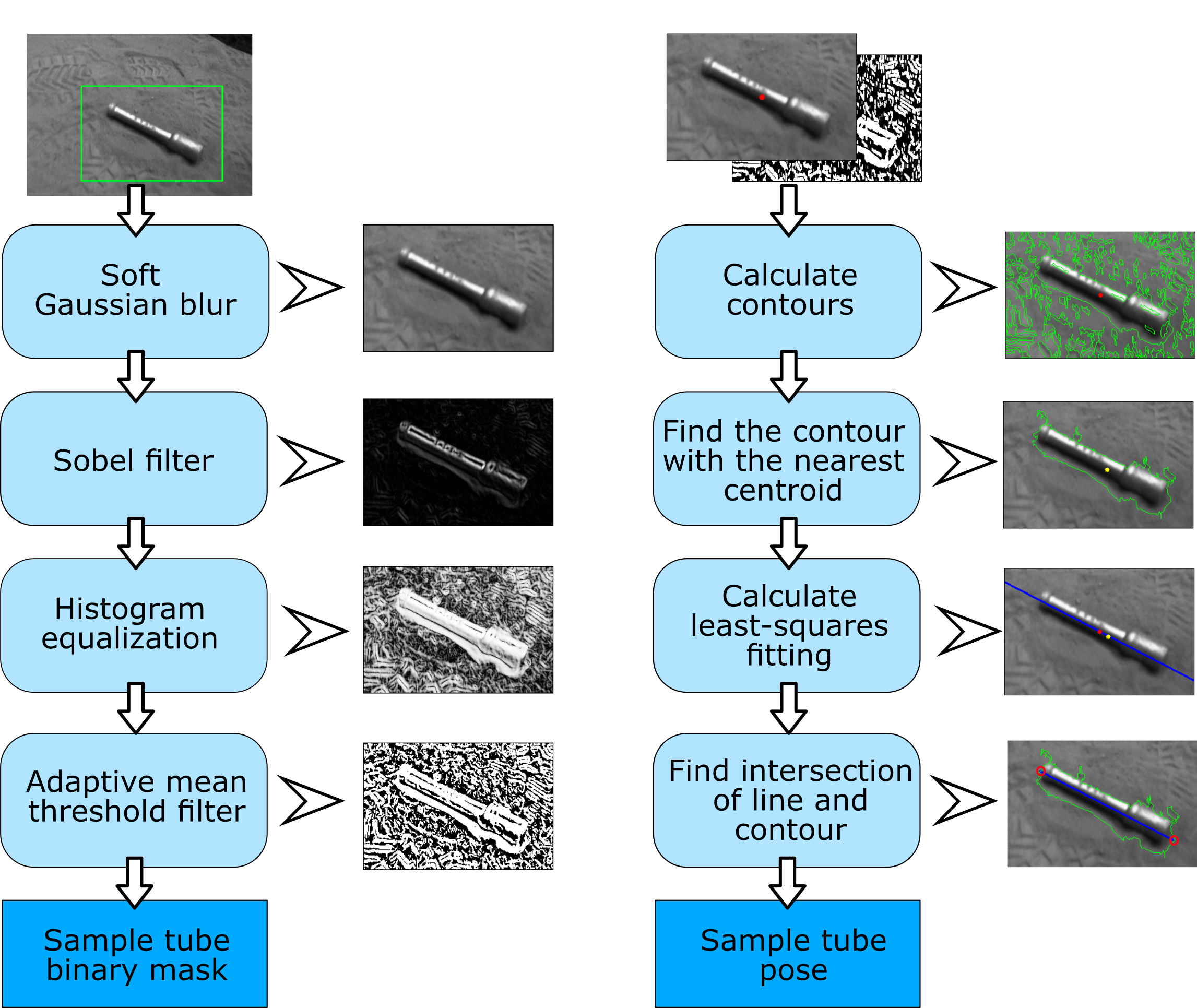}
\caption{Diagram of the algorithms employed to estimate the sample tube pose using CV techniques. On the left, a binary mask is produced when the sample tube is detected. On the right, the sample tube pose is calculated using the binary mask.}
\label{fig:pose-estimation}
\end{figure}

The pose estimation module process is divided into two steps (see Fig.~\ref{fig:pose-estimation}). In the first one (left side of figure), the cropped area determined by the bounding box is employed to calculate the binary mask of the sample tube. The computational load is lowered, as only a reduced portion of the image is processed. First, a soft Gaussian blur is performed to reduce image imperfections, followed by applying a Sobel filter to obtain the polygonal intensity gradients of the sample tube edges. The resultant image histogram is equalized to enhance the acquired gradients. Finally, an adaptive mean threshold filter is applied to produce the required binary mask.

In the algorithm depicted on the right side of Fig.~\ref{fig:pose-estimation}, the sample tube centroid and its previous binary mask are employed to obtain the final sample tube pose. First, all the mask contours are computed along their moments. A first screening is performed to choose those contours which enclose the centroid. Afterwards, a second screening takes place, focused on locating the contour whose distance is minimal to the centroid. The resulting contour is chosen as the one belonging to the sample tube. Once the sample tube contour is found, the algorithm proceeds to find its orientation: a least-squares fitting is performed on all the image points enclosed by the contour. Like this, the slope that represent the sample tube orientation is obtained. To correct possible errors introduced in the generation of the contour, a line is created using the slope and the sample tube detection centroid. Ultimately, the sample tube end points are calculated as a result of the intersection between the contour and the line. These points can be mapped to real world coordinates with the help of the disparity maps provided by the stereo cameras. Thus, the Pinhole Camera Model is employed to establish a relationship between the 2D points of the disparity maps and the real world 3D coordinates, which is finally output by the module. 

\begin{figure*}[t]
    \centering
    \subfloat[]{
    \includegraphics[width=0.47\textwidth]{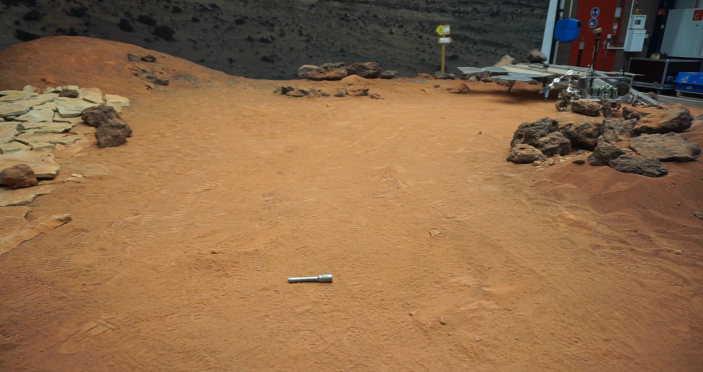}
    }
    \quad
    \subfloat[]{
    \includegraphics[width=0.47\textwidth]{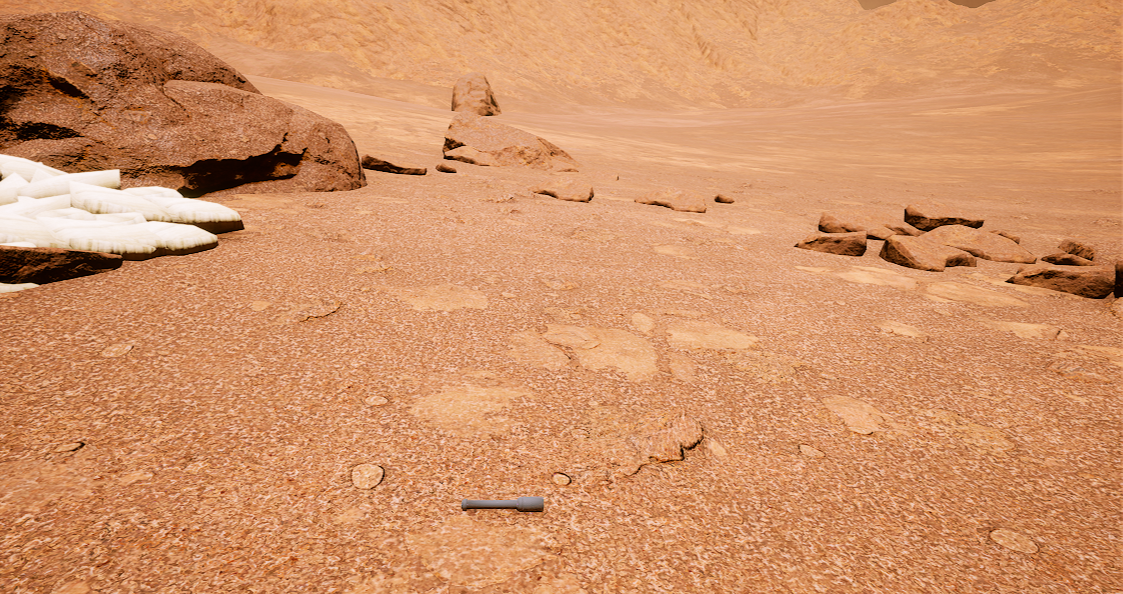}
    }
    \caption{Figure (a) shows the sample tube on sandy terrain in the ESA's PUTB. Rocks of several sizes are observed on the background. Picture (b) depicts the sample tube in an Unreal Engine 4 simulation. Rocks and obstacles produce shadows depending on a virtual sun position. }\label{fig:unreal-sample}
\end{figure*}

\section{EXPERIMENTAL RESULTS} 
To validate the proposed Sample Localization system, a dataset with real and artificial images was generated. It was later used to evaluate the sample detection and pose estimation through the proposed DNN and CV techniques. Finally, the system was deployed on the ExoTeR testbed rover to demonstrate its ability to retrieve sample tubes in a laboratory scenario.

\subsection{Mars sample localization dataset}
The objective of this dataset was to provide enough virtual and real images to validate the proposed system. 
Two environments were used for the training dataset generation.
On the one hand, synthetic images were generated using Unreal Engine 4, a videogame engine with a focus on visual realism. A Martian environment was recreated, depicting the sample tube in different illumination conditions. Additionally, UnrealCV plugin \cite{Qiu2017} was used to take in-game images. A picture of the sample tube in a Martian scenario is shown in Fig. \ref{fig:unreal-sample}b.
On the other hand, real images of the sample tube were taken at the Planetary Utilisation Testbed (PUTB) located at ESA. It consists on a 9 $m^{2}$ square Mars-like tesbed filled with different terrains and rocks. 
For this purpose, the ExoTeR \cite{azkarate2022design} rover was used, depicted in Fig.~\ref{fig:exoter-rover}. The rover contains an Intel Core i7-7600 processor 2.80 GHz and 16GB of RAM and is equipped with two Bumblebee2 stereo vision cameras: LocCam and NavCam. The cameras produce grayscale images with a resolution of $1024\times768$ pixels per image sensor and a focal length of \SI{3.8}{\mm} (\SI{66}{\degree} HFOV). 
An image of the sample tube on the terrain from the NavCam point of view is depicted in Fig.~\ref{fig:unreal-sample}a.

The global dataset has been published in Zenodo\footnote{\url{http://doi.org/10.5281/zenodo.6542933}}. It has been divided into three datasets of images. 
The first one related to the generation of images for training purposes, which is composed of 602 synthetic and 52 real images.
The second dataset include 316 and 84 real images taken by the NavCam and LocCam respectively and processed to show only the sample tube bounding box. These images also include the sample tube centroid (obtained from the DNN), and were used to validate the sample pose estimation module.
Finally, the third dataset includes 2178 images that were taken during the laboratory tests with the ExoTeR rover performing two sample retrieval tasks. 

\subsection{DNN Implementation}
The YOLOv3-tiny DNN model was trained using its framework based on Darknet architecture. The source code was published under MIT open source license\footnote{\url{https://github.com/spaceuma/MarsSampleLocalization}}.
In order to emulate a low availability of real images, the approach presented in Section~\ref{sec:detection} was trained using the corresponding dataset, which was composed of much more synthetic images than real ones.

The provided trained model was later implemented into the Google's Coral USB accelerator to perform the sample tube detection. This hardware-accelerator is composed of an edge Tensor Processing Unit (edge TPU), which is an ASIC specialized in the inference of neural networks. Furthermore, it provides an excellent compromise between performance and electrical consumption \cite{libutti2020}. The TPU was connected to an ARM Cortex-A72 processor (\SI{1.5}{\GHz} and \SI{4}{GB} of RAM) as a CPU that could be available in further space missions \cite{powell2018high}. 
Some operational transformations were applied to the network to embed it into the hardware-accelerator. Firstly, a process of quantization was followed in which the default 32-bit data of the network tensor was transformed to a 8-bit fixed-point representation. Secondly, some mathematical operations were changed to equivalent ones supported by the TPU. For example, the leaky Rectified Linear Unit (ReLU) activation layers were replaced by standard ReLu layers.
Furthermore, it was necessary to take into account that the YOLOv3-tiny model works with squared images of $416\times416$ pixels of resolution as inputs. Thus, larger images, as the ones obtained from the cameras, were scaled, keeping the aspect ratio using an operation of zero-padding. This, together with the quantization process, helped to reduce computation costs.
The trained model was also implemented on a general purpose computer with an Intel Core i7 CPU (\SI{2.60}{\GHz} and \SI{16}{GB} of RAM) and a NVIDIA GeForce RTX 2070 GPU to compare its performance over the hardware accelerator.
A comparison between different hardware configurations of both YOLOv3 and YOLOV3-tiny networks was carried out to evaluate the inference of the proposed Sample Localization approach. For this purpose, the laboratory test dataset was used and the obtained results are shown in Table~\ref{tab:inference-times}.
Overall, the proposed YOLOv3-tiny model provides faster inference times than the standalone YOLOv3 version. The fastest time were achieved using the Intel Core i7 + GPU configuration. This was expected, as GPUs are devices traditionally used to run neural networks due to their high performance through high power consumption. Second best inference times were obtained by the Intel Core i7 + TPU configuration, closely followed by the ARM + TPU setup. The slowest times were produced when the DNN was running on a CPU. Based on the obtained information, the inference time of the YOLOV3-tiny model on an ARM CPU + TPU was 84 times faster than without TPU, providing lower power consumption and size.
\begin{figure}
  \centering
  \includegraphics[width=0.48\textwidth]{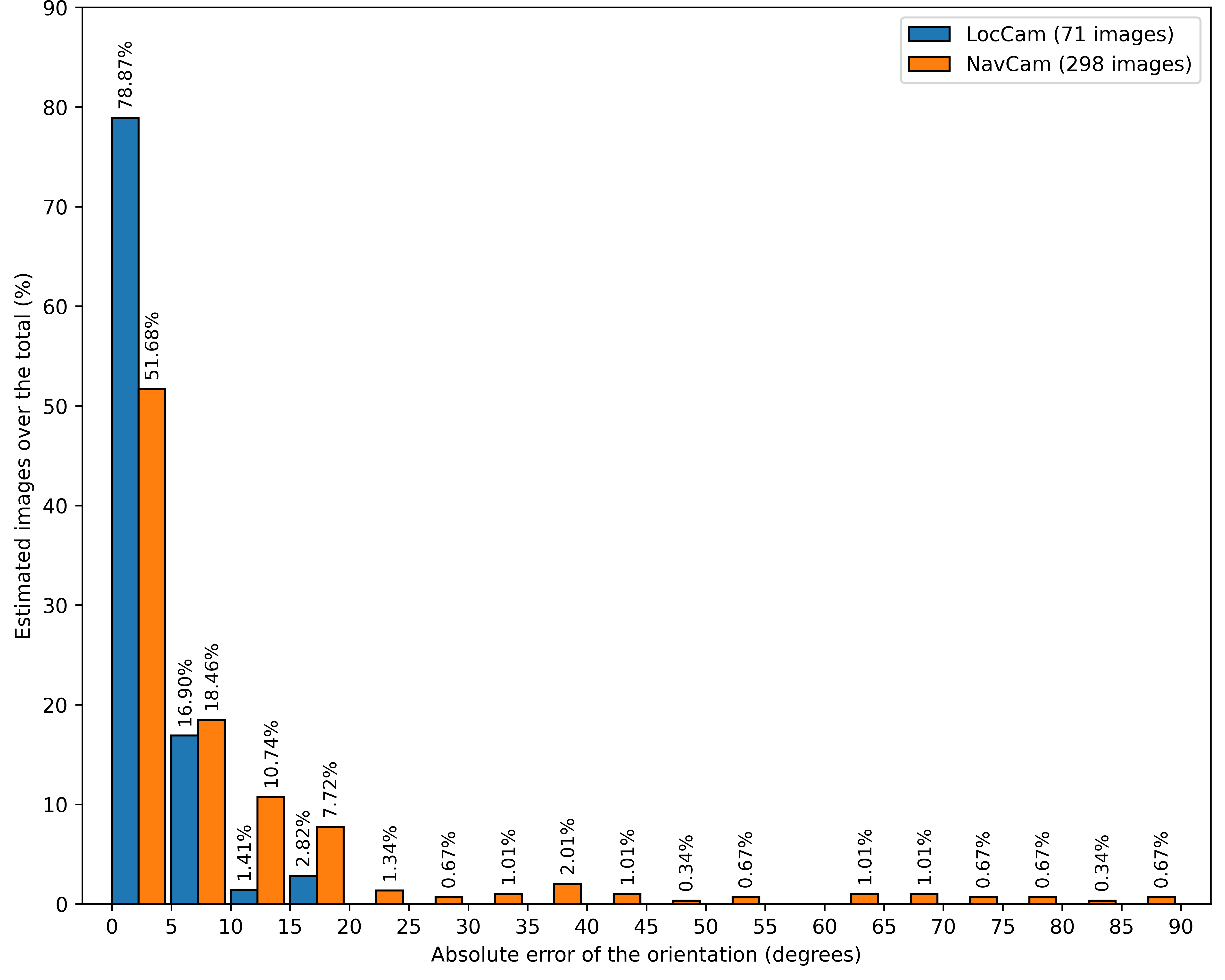}
\caption{Charts with the errors obtained in the estimation of sample tubes orientation in LocCam (blue bars) and Navcam (orange bars) images. Errors of \SI{90}{\degree} are considered the maximum possible, since the estimated orientation would be perpendicular to that of the samples.}
\label{fig:orientation-errors}
\end{figure}
\begin{table}
\centering

\caption{Inference times (ms) of YOLOv3-tiny and YOLOv3 on different platforms}
\begin{tabular}{lcc}
\hline
Device & YOLOv3-tiny & YOLOv3 \\
\hline
Intel Core i7 & 140.20 &   1292.72 \\
Intel Core i7 + GPU & 5.12  & 18.79  \\
Intel Core i7 + TPU & 21.03 &  152.04   \\
ARM Cortex-A72 & 2808.62 &  52787.52  \\
ARM Cortex-A72 + TPU & 33.46  & 222.54  \\
\hline
\end{tabular}
\label{tab:inference-times}
\end{table}

\subsection{Sample pose estimation}
The sample pose estimation module was validated using the second dataset, where sample tubes images with its centroid were generated. 
Results of the orientation estimation can be seen in Fig.~\ref{fig:orientation-errors}. For the LocCam, the algorithm presented an average error of $3.23\pm3.38$º while for the NavCam, it presented an average error of $10.75\pm16.47$º. According to the chart, all the LocCam estimations had an error less than 20º, with most of the estimations (\SI{78.87}{\percent}) having an error less than \SI{5}{\degree}.
This is considered acceptable since the pose estimation is required when the rover is just in front of the sample tube, i.e., using LocCam images. It is assumed the manipulator gripper is able to pick up a sample tube with an orientation error lower than \SI{30}{\degree}. In the NavCam case, most of the estimations (\SI{51.68}{\percent}) had an error of less than \SI{5}{\degree} but presented greater errors due to the distance of the sample tube. However, it does not pose a problem since the NavCam is not used to estimate the sample orientation and command the manipulator.
As for the computation times employed for the CPUs to provide an estimation of the sample tube pose, the Intel Core i7 lasted 1.94 ms while the ARM Cortex-A72 took 14.12 ms. 

\subsection{ExoTeR laboratory tests}

\begin{figure}
  \centering
  \includegraphics[width=0.48\textwidth]{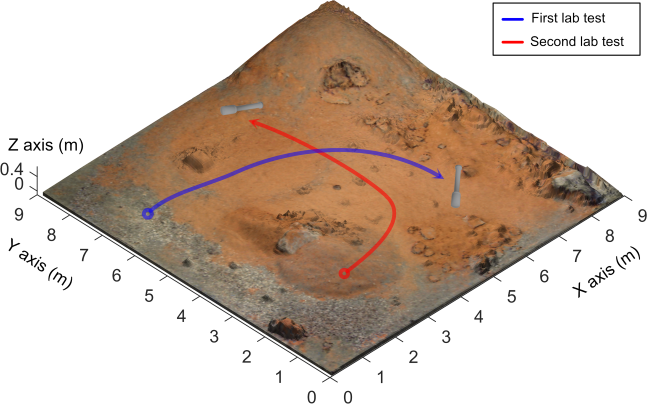}
\caption{Trajectories followed by ExoTeR in the PUTB on both laboratory tests to reach the sample tube. In both of them, the rover starts from a position in which the sample tube is not visible.}
\label{fig:prl-traj}
\end{figure}

\begin{figure}
  \centering
  \subfloat[]{
  \includegraphics[height=0.13\textheight]{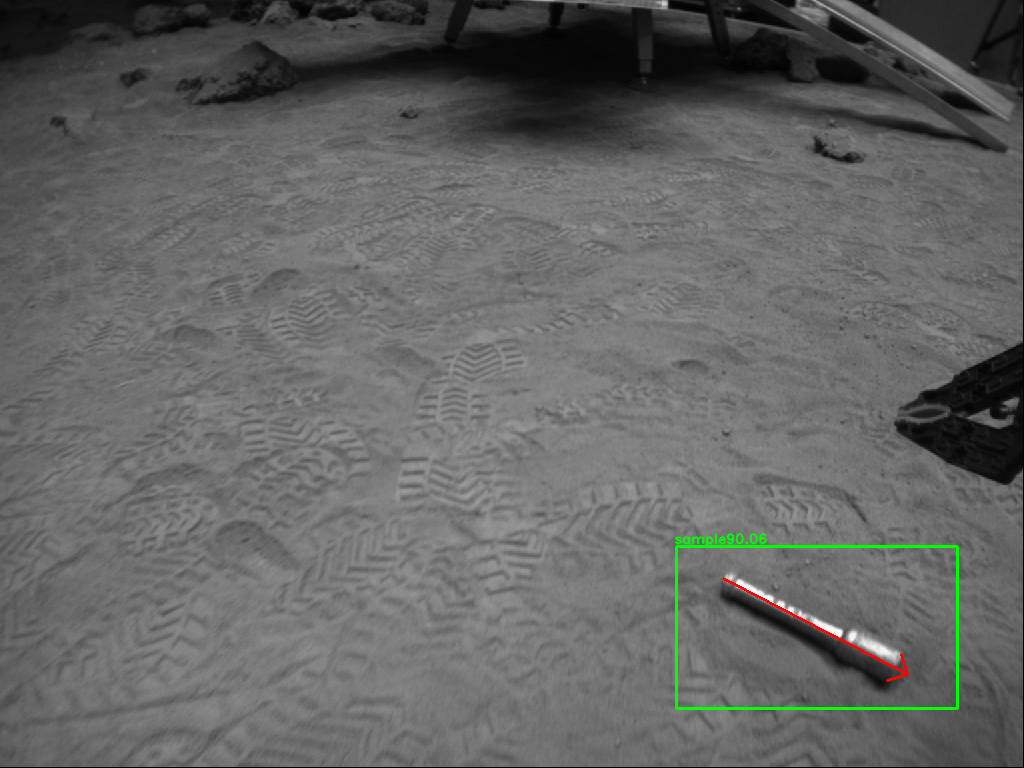}
  }
  \subfloat[]{
  \includegraphics[height=0.13\textheight]{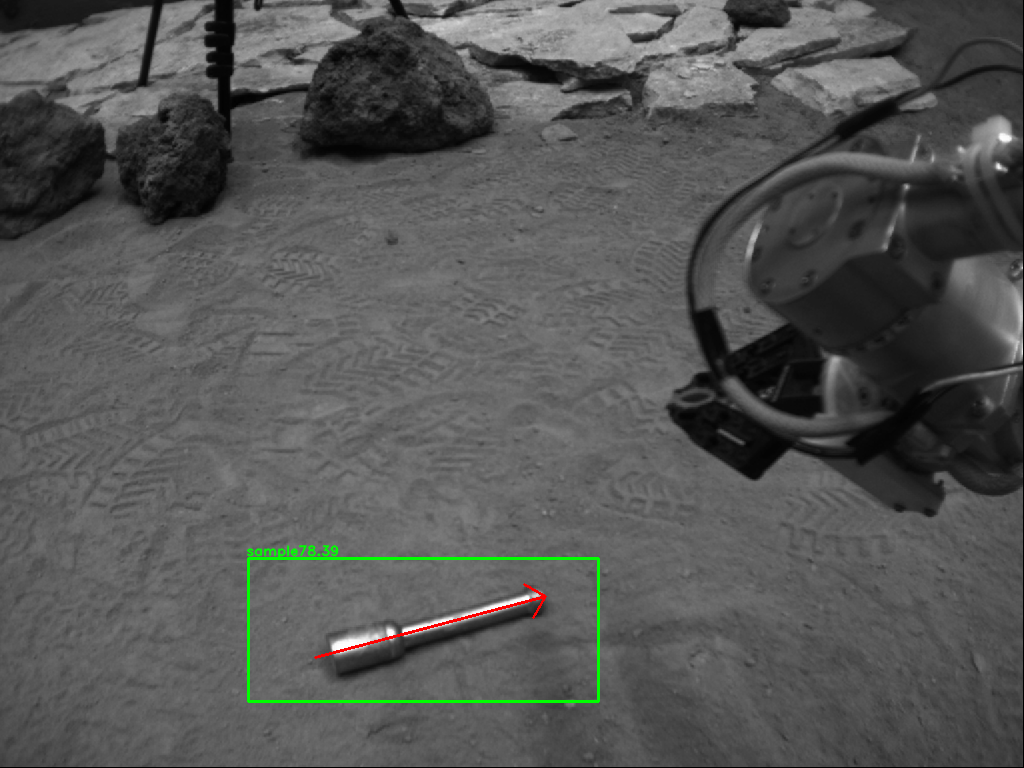}
  }
\caption{Detection bounding box (green rectangle) and confidence score along pose estimation (red arrow) of lab test 1 (a) and lab test 2 (b) images obtained with the LocCam.}
\label{fig:final-det}
\end{figure}

\begin{figure}
  \centering
  \includegraphics[width=0.48\textwidth]{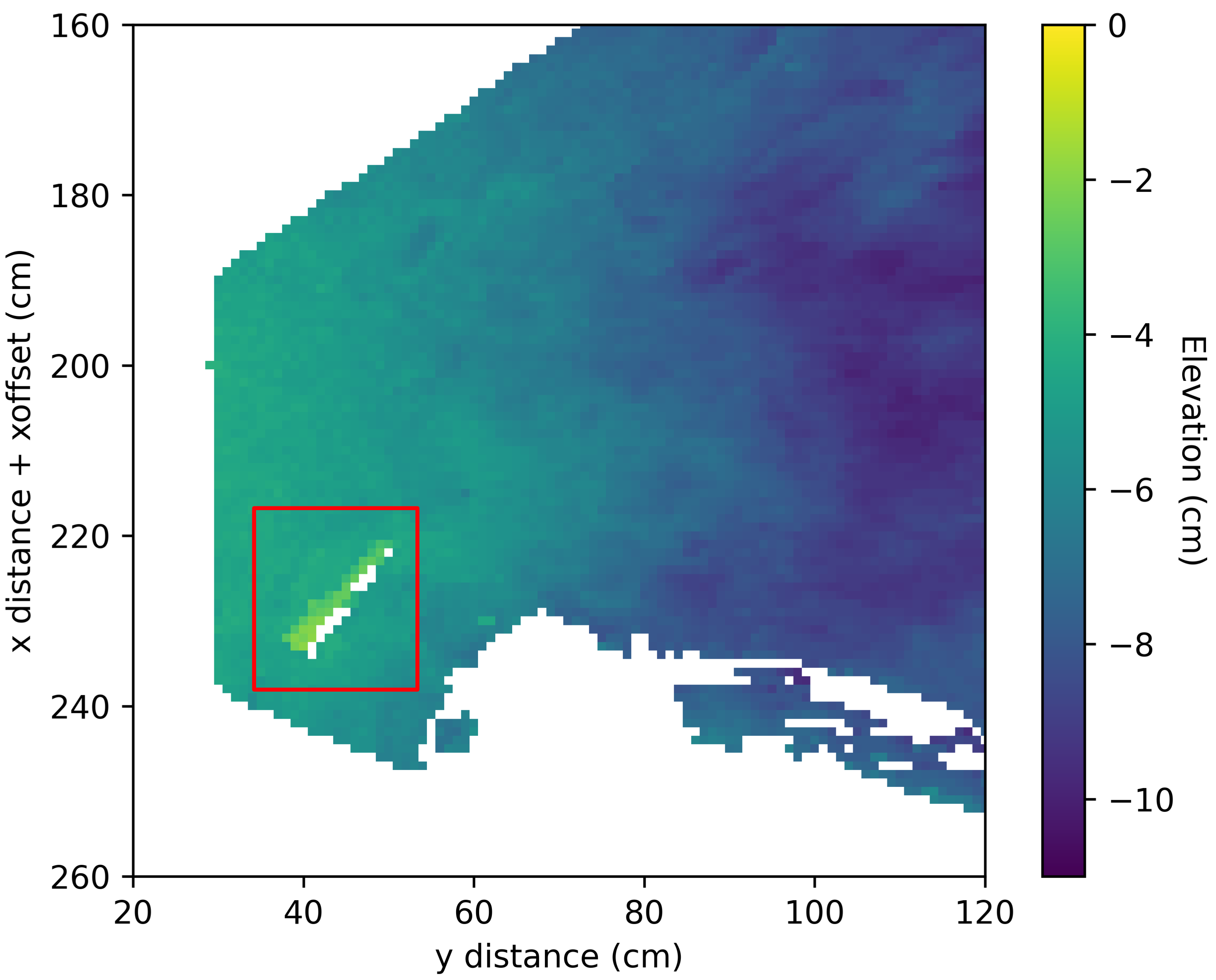}
\caption{DEM generated from LocCam stereo images using the Pinhole Camera Model. This map corresponds to the final step of laboratory test 1. The sample tube is located inside the red box. }
\label{fig:DEM-map}
\end{figure}

Two laboratory tests were carried out using the ExoTeR rover on the PUTB. The objective was to validate the proposed system and demonstrate it is suitable to perform sample tube retrievals.
For this purpose, the rover had to reach a sample tube located on the terrain to pick it up with its manipulator. At the beginning of the tests, the sample tube position was provided outside the rover field of view, and approximate coordinates were given to the Guidance, Navigation and Control (GNC) subsystem of the rover. This stage simulated the orbital coordinates that would be provided during the SFR mission. It is worth highlighting the rover was not provided with any external ground truth device, e.g., Vicons, GNSS. Instead, a visual odometry algorithm was used \cite{Geiger2011StereoScan:Real-time}. Once the rover started moving, the rover goal trajectory was automatically corrected with the sample coordinates derived from the Sample Localization system. Once it was reachable, the sample tube pose was computed to finally place the rover with the right orientation to pick up the sample. The trajectories followed by ExoTeR are represented in Fig.~\ref{fig:prl-traj}.
A video of these two experiments was recorded and published\footnote{\url{https://youtu.be/8_ymP6bg6-c}}.
Images from both stereo cameras (NavCam and LocCam) were continuously fed to the neural network for the sample tube detection. This redundancy contributed to the robustness of the process, as each image provided a different point of view of the scene. This approach was possible due to the low latency of the inference, which was done by the use of the onboard CPU+TPU configuration. In the final step, the sample tube world coordinates and orientation were obtained using the LocCam disparity map. An example of this step can be seen in Fig.~\ref{fig:final-det} where sample tubes are detected along their pose. Additionally, Fig.~\ref{fig:DEM-map} shows the obtained Digital Elevation Map (DEM) in which the sample tube is present.

These two lab tests provided information about the detection using the pre-trained model with synthetic and real training images. During the tests, the system was able to detect the sample tube up to 5 m of distance from the rover using the Navcam. According to the cameras FoV and this maximum distance, it can be assumed the rover will be able to detect samples with a localization error up to 1.5 m. Table~\ref{tab:performance-metrics} shows the obtained performance in both YOLOV3 and YOLOv3-tiny models deployed on a PC and the TPU, where a score confidence threshold of 0.75 was used to consider the sample was detected. Both networks were trained for 2000 iterations, choosing the weights with best detection rate on the validation set. No comparison between the detection rate of YOLOv3-tiny and YOLOv3 will be made as different parameters can be tested to increase it and the objective of the experiment is to verify the quality of the hardware acceleration and the transfer learning procedure. In addition, the detections of the networks trained with only real images are also presented. 
As can be observed in the YOLOv3-tiny network data, the pretrained model produced an increment of true positives of \SI{56}{\percent} and \SI{26}{\percent} respectively on both laboratory tests over the real version. The network robustness to false positives was enhanced thanks to the transfer learning approach, achieving no false positives in any of the pretrained models. Comparing the YOLOv3-tiny TPU implementation with its PC counterpart, there is no significant decrease of the detection rate. As for the standalone YOLOv3 model, a significant degradation on its detection rate is observed. This is mainly due to its high number of layers, that needs to be ported to the hardware-accelerator, losing accuracy in the process, considering YOLOv3-tiny more suitable to be deployed on the TPU.

\begin{table}[h]
\centering
\caption{Laboratory tests detection metrics of YOLOv3-tiny and YOLOv3 implemented in the TPU and a PC}
\begin{tabular}[t]{llcccc}
& & \multicolumn{2}{c}{Lab test 1} &  \multicolumn{2}{c}{Lab test 2} \\ \cline{3-6} 
& & Real & Pretrained & Real & Pretrained  \\
\hline
\multirow{3}{1.24cm}{YOLOv3-tiny (TPU)} 
& True Positives     & 209  & 323  & 774  & 978 \\
& False Negatives    & 595  & 482  & 494  & 318 \\
& False Positives    & 0    & 0    & 3     & 0  \\
\hline
\multirow{3}{1.24cm}{YOLOv3-tiny (PC)} 
& True Positives     & 277  & 363  & 933  & 960 \\
& False Negatives    & 536  & 440  & 467  & 445 \\
& False Positives    & 0    & 0    & 4     & 0  \\
\hline
\multirow{3}{1.24cm}{YOLOv3 (TPU)} 
& True Positives     & 5    & 15   & 5    & 3    \\
& False Negatives    & 795  & 785  & 1255 & 1257 \\
& False Positives    & 0    & 0    & 0    & 0    \\
\hline
\multirow{3}{1.24cm}{YOLOv3 (PC)} 
& True Positives     & 144  & 149   & 480    & 768    \\
& False Negatives    & 658  & 653  & 1025   & 575 \\
& False Positives    & 0    & 0     & 0    &  0  \\
\hline
\end{tabular}
\label{tab:performance-metrics}
\end{table}

\section{CONCLUSIONS}

This paper proposes a system architecture to detect the pose of a sample that would be picked up by a robot. It has been used to perform an experiment using a rover testbed from ESA equipped with a manipulator.
Obtained results demonstrate the use of a hardware accelerator could improve the required processing time to detect a sample tube, as well as it reduces the CPU load. Moreover, the use of synthetic images improved the detection of sample tubes without requiring large datasets of real images. Finally, the use of CV to compute the sample tube position and orientation has been demonstrated to be accurate and fast enough for the performed grasping operations, reducing the DNN architecture complexity.
Additionally, this architecture can be extrapolated to other robotic platforms, such as drones as recently announced by NASA, with few improvements: the use of different point of views, i.e. ground vs aerial images, could be solved by increasing the training dataset with this kind of images; and the use of RGB cameras could improve the detection since they can provide more feature information to the DNN.
Future work will be focused on the enhancement of the independent subsystems as well as to perform experiments of the system on sample fetching aerial robots. For the detection subsystem, studies will be centered on the implementation of different topologies of neural networks on different hardware accelerators, such as other DNN devices or FPGAs as a more space-representative hardware. On the other hand, the pose estimation subsystem will be improved by incorporating 3D information to analyse its performance with different objects, i.e. not only sample tubes, comparing it to the proposed approach in this paper.

\bibliographystyle{IEEEtran}
\bibliography{references}

\vfill
\end{document}